\documentclass{article}

\usepackage{microtype}
\usepackage{graphicx}
\usepackage{booktabs} 
\usepackage{amsmath}
\usepackage{amssymb}

\usepackage{pifont,wasysym,multirow}
\usepackage{makecell}

\usepackage{xcolor}

\newcommand{\yes}{\ding{51}}
\newcommand{\no}{\ding{55}}
\newcommand{\half}{$\LEFTcircle$}
\newcommand{\na}{--}

\newcommand{\dingone}{\ding{172}}
\newcommand{\dingtwo}{\ding{173}}

\newcommand{\dingOne}{\ding{182}}
\newcommand{\dingTwo}{\ding{183}}
\newcommand{\dingThree}{\ding{184}}
\newcommand{\dingFour}{\ding{185}}
\newcommand{\dingFive}{\ding{186}}
\newcommand{\dingSix}{\ding{187}}

\usepackage{tcolorbox}
\tcbuselibrary{breakable,listings,skins}
\definecolor{questionaccent}{HTML}{29465B}
\definecolor{questionbackground}{HTML}{F5F7F9}
\newtcblisting{generalquestions}{enhanced,breakable,listing only,colback=questionbackground,colframe=questionaccent!35,colbacktitle=questionaccent!8,coltitle=questionaccent,fonttitle=\small\sffamily\bfseries,title={General-knowledge questions},title after break={General-knowledge questions (continued)},boxrule=0.4pt,arc=1mm,left=7mm,right=3mm,top=2mm,bottom=2mm,before skip=8pt,after skip=8pt,listing options={language={},basicstyle=\small\ttfamily,columns=fullflexible,keepspaces=true,breaklines=true,breakatwhitespace=true,showstringspaces=false,numbers=left,numberstyle=\scriptsize\sffamily\color{questionaccent},numbersep=8pt,aboveskip=0pt,belowskip=0pt}}

\usepackage{hyperref}
\usepackage{xurl}

\usepackage[accepted]{mlsys2025}
\hypersetup{pdfsubject={}}
\makeatletter
\renewcommand{\Notice@String}{}
\makeatother

\makeatletter
\let\mlsys@makecaption\@makecaption
\long\def\@makecaption#1#2{\def\@tempa{figure}\ifx\@captype\@tempa\vskip -7pt\fi\mlsys@makecaption{#1}{#2}}
\makeatother

\mlsystitlerunning{CertiSum: Bitwise-Identical MXFP4 Training Across Platforms and Parallelism Layouts}

\begin{document}
\setlength{\footskip}{24pt}
\fancyfoot[C]{\small\thepage}
\twocolumn[
\mlsystitle{CertiSum: Bitwise-Identical MXFP4 Training\texorpdfstring{\\}{ }Across Platforms and Parallelism Layouts}



\mlsyssetsymbol{equal}{*}

\begin{mlsysauthorlist}
\mlsysauthor{Steve Rhyner}{1,2}
\mlsysauthor{Yubo Gao}{1,2,3}
\mlsysauthor{Stefan de Lasa}{1,2}
\mlsysauthor{Sankeerth Durvasula}{1,2}
\mlsysauthor{Nandita Vijaykumar}{1,2}
\end{mlsysauthorlist}

\mlsysaffiliation{1}{University of Toronto}
\mlsysaffiliation{2}{Vector Institute}
\mlsysaffiliation{3}{NVIDIA}

\mlsyscorrespondingauthor{Steve Rhyner}{rhyners@cs.utoronto.ca}

\mlsyskeywords{}

\vskip 0.3in

\begin{abstract}
Training with MXFP4, a 4-bit microscaling format, is non-deterministic: identical inputs can produce different results across parallelism layouts and GPU vendors. These differences complicate debugging and training audits and can cause training and inference engines to disagree. We present CertiSum to make MXFP4 training bitwise invariant across runs, parallelism layouts, and GPU vendors. Its key insight is that native FP32 accumulation of MXFP4 products is often already exact for every summation order. CertiSum collects statistics during quantization to certify native matrix multiplication results, recomputes uncertified outputs exactly, and uses exact kernels directly where certification rarely succeeds. CertiSum keeps quantization blocks unchanged across layouts and merges partial sums without rounding, including in attention and weight-gradient computation. In 200-step Qwen3-8B-Base experiments, CertiSum preserves training states bit for bit across AMD MI350X and NVIDIA GB200 and across changes in data, tensor, and context parallelism on MI350X. Across three 8B--14B models on FineWeb-Edu using MI350X, it adds 11.8--60.5\% to training-step latency relative to its non-exact MXFP4 counterpart at the larger evaluated input shapes, with speedups of up to $1.24\times$ over BF16 and $1.70\times$ over deterministic BF16.
\end{abstract}
]



\printAffiliationsAndNotice{}  

\section{Introduction \& Motivation}
\label{sec:intro}
The growing scale of large language models (LLMs) has made compute, memory, and energy major costs of training, motivating narrower numerical formats~\cite{brown2020language,gholami2024ai,abecassis2025pretraining}. Recent developments in low-precision formats have allowed 4-bit floating-point (FP4) and its corresponding microscaling (MX) format, MXFP4, to serve production models such as OpenAI’s gpt-oss~\cite{agarwal2025gpt}. Compared to FP8, FP4 halves memory use and at least doubles matrix throughput on modern accelerators~\cite{abecassis2025pretraining}. With the correct training recipe, MXFP4 has been shown to achieve model quality close to BF16 at a significantly lower estimated cost of the backward pass~\cite{tseng2025training}.

\begin{figure}[!t]
    \centering
    \includegraphics[width=\linewidth]{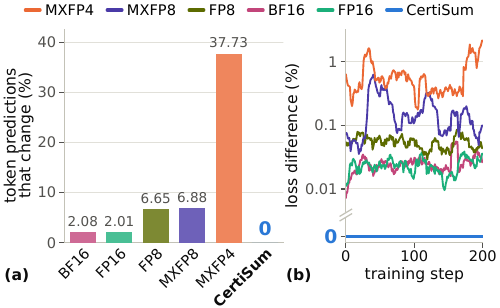}
    \caption{Qwen3-8B-Base~\cite{yang2025qwen3} on FineWeb-Edu~\cite{penedo2024fineweb}, trained on 4 GPUs instead of 2, all else identical. (a) Differing greedy next-token predictions before training. (b) Relative training-loss difference (centered 10-step mean). Repeating the run of each format on two GPUs reproduces the result bit for bit; CertiSum (\S\ref{sec:CertiSum}) differs in neither predictions nor loss.}
    \label{fig:motivating_example}
\end{figure}

Like all floating-point (FP) training, low-precision training is affected by non-determinism. Reductions in machine learning (ML) frameworks can return bitwise-different outputs for the same inputs, for two reasons. First, floating-point addition is non-associative, so the result depends on the summation order, which is set by how each reduction is split across GPUs by data, tensor, and context parallelism (DP, TP, CP)~\cite{li2020pytorch,shoeybi2019megatron,liu2024ringattention} and across streaming multiprocessors (SMs) within each kernel. We call this partitioning the \emph{parallelism layout} (\S\ref{sec:parallelism_layout}). Atomic additions, common in the attention backward pass~\cite{qiang2026dash,shah2024flashattention}, further make the order depend on thread scheduling. Second, the kernel that performs a given operation is itself not fixed. Autotuners and framework heuristics select among implementations based on shape, hardware, and timing measurements, and different implementations reduce in different orders.

Non-determinism undermines several common ML workflows. In reinforcement learning (RL), an inference engine samples tokens and a training engine independently computes their log-probabilities. The two engines often use different kernels and parallelism layouts~\cite{sheng2025hybridflow,chen2025minimax}, so nominally on-policy training becomes off-policy~\cite{yao2025offpolicy}. Debugging a training run requires reproducing a step with its exact parallelism layout and kernel selection. Without determinism, discrepancies caused by non-deterministic execution are hard to distinguish from implementation errors~\cite{aithal2026megatrondeterminism}, such as a numerically unstable algorithm (e.g., a na\"ive exponential in softmax). Non-determinism also weakens kernel testing. A kernel and its reference implementation sum in different orders, so tests must compare them within a tolerance, and such tolerances could accept many incorrect kernels~\cite{sarkar2026correctness,shah2026contract}. Training audits, which re-execute a published training run, possibly on different accelerators, to verify that the released weights follow from the claimed data and code, rely on reproducibility as well. Without it, a backdoor in the model may go undetected~\cite{donaghy2026open1bfullyauditabletraining}.

MXFP4 training is more sensitive to the parallelism layout than BF16 training. Figure \ref{fig:motivating_example} measures how an LLM's output tokens and training loss change when training partitions the model across 4 GPUs instead of 2 (using tensor parallelism). Under MXFP4, $18\times$ more output tokens change than under BF16. The workflows above are therefore most affected in the low-precision formats that make training cheaper.

To address this, we propose CertiSum, a method that provides \textit{invariance}, the property that the values an LLM workload produces depend only on the model and its inputs, not on how or where the computation executes. These values are the model outputs during inference and the losses, gradients, parameters, and optimizer states during training. We target three forms of invariance.
Values must be identical across repeated runs with the same configuration (\textit{run-to-run invariance}), across parallelism layouts and micro-batch counts (\textit{parallelization invariance}), and across accelerators from different vendors (\textit{hardware invariance}). Kernels may use any reduction strategy that yields invariant results.
We aim to provide these invariances at low latency overhead relative to non-deterministic execution. At production scale, even small overheads are costly~\cite{grattafiori2024llama, meta2024llama31card}.

Prior work provides only a subset of these invariances, or provides all of them at high cost (Table~\ref{tab:related}). Order-fixing systems~\cite{aithal2026megatrondeterminism,he2025nondeterminism,zhang2025deterministic,jiang2026isoexec,li2023easyscale} provide run-to-run invariance but no hardware invariance across vendors. Most provide parallelization invariance only for specific layout changes (batch size, TP, or DP degree), and some only in the forward pass. Cross-hardware replay~\cite{donaghy2026open1bfullyauditabletraining} provides hardware invariance at 5.2--6.8$\times$ the runtime of BF16 training. Order-free summation methods~\cite{ahrens2020algorithms,collange2015numerical,iakymchuk2016exblas} provide all three invariances but with high overhead (up to 16$\times$ slower than a non-reproducible general matrix multiplication (GEMM)~\cite{iakymchuk2016exblas}), since they pay for exactness on every reduction. FoldAttention~\cite{achanta2026foldattention} avoids this overhead but covers only attention. Exact and emulated GEMMs either guarantee exactness only up to a reduction length fixed before runtime~\cite{hayashi2026awe} or target FP64 accuracy rather than invariance~\cite{schwarz2026guaranteed}. No existing system provides all three invariances for MXFP4 training across DP, TP, and CP with low overhead.

\begin{table*}[t]
\centering
\footnotesize
\caption{Comparison with prior work. A \yes\ means guaranteed or supported, \half\ partial or argued only, \no\ not provided or out of scope, -- not addressed, and n/r not reported. Overhead is latency relative to non-deterministic execution.
\textsuperscript{a}~NVIDIA GPUs only.
\textsuperscript{b}~236\% across GPU types (CNNs).
\textsuperscript{c}~Claimed for any device count; no replay on a different device count is reported.
\textsuperscript{d}~CPU only.
\textsuperscript{e}~Latency overhead of 1.2-1.6$\times$ argued from previous work.
\textsuperscript{f}~GEMM latency overhead~\cite{iakymchuk2016exblas}.
\textsuperscript{g}~Attention only.
\textsuperscript{h}~Targets accuracy, not reproducibility.}
\vspace{4pt}
\setlength{\tabcolsep}{1.5pt}
\begin{tabular}{l l c cccc c c c l l}
\toprule
 & & Run-to- & \multicolumn{4}{c}{Parallelization} & Hard- & Order- & Back- & Overhead & Precision \\
\cmidrule(lr){4-7}
 & & run & Batch & \makebox[2em]{DP} & \makebox[2em]{TP} & \makebox[2em]{CP} & ware & free & ward & (reported) & (evaluated) \\
\midrule
\multirow{5}{*}{\shortstack[l]{\emph{Order-}\\\emph{fixing}}}
& Megatron Core~\cite{aithal2026megatrondeterminism} & \yes & \no & \no & \no & \no & \no & \no & \yes & 2\% & FP8, FP4 \\
& \shortstack[l]{Batch-inv.\ kernels\\\cite{he2025nondeterminism}} & \yes & \yes & \na & \na & \na & \na & \no & \no & 1.6--2.1$\times$ & BF16 \\
& TBIK~\cite{zhang2025deterministic} & \yes & \yes & \half & \yes & \na & \na & \no & \na & 22--63\% & BF16 \\
& IsoExec~\cite{jiang2026isoexec} & \half & \yes & \half & \yes & \no & \na & \no & \na & 25\% & n/r \\
& EasyScale~\cite{li2023easyscale} & \yes & \no & \yes & \na & \na & \half\textsuperscript{a} & \no & \yes & $<$1\%\textsuperscript{b} & n/r \\
\midrule
\multirow{2}{*}{\shortstack[l]{\emph{Cross-}\\\emph{hardware}}}
& OPEN-1B~\cite{donaghy2026open1bfullyauditabletraining} & \yes & \na & \half\textsuperscript{c} & \na & \na & \yes & \no & \yes & 5.2--6.8$\times$ & INT8 + BF16 \\
& RepDL~\cite{xie2025repdl} & \half & \half & \na & \na & \na & \half & \no & \half & n/r & FP32 \\
\midrule
\multirow{3}{*}{\shortstack[l]{\emph{Order-}\\\emph{free}}}
& ReproBLAS~\cite{ahrens2020algorithms} & \yes & \half & \half & \half & \half & \half\textsuperscript{d} & \yes & \na & n/r\textsuperscript{e} & FP32, FP64 \\
& ExBLAS~\cite{collange2015numerical} & \yes & \half & \half & \half & \half & \yes & \yes & \na & $\leq$16$\times$\textsuperscript{f} & FP64 \\
& FoldAttention\textsuperscript{g}~\cite{achanta2026foldattention} & \yes & \yes & \na & \na & \na & \na & \half & \yes & $\approx$0\% & BF16 \\
\midrule
\multirow{2}{*}{\shortstack[l]{\emph{Exact}\\\emph{GEMM}}}
& AWE~\cite{hayashi2026awe} & \half & \na & \na & \na & \na & \half & \yes & \na & n/r & INT8 on FP4 \\
& ADP~\cite{schwarz2026guaranteed} & \na & \na & \na & \na & \na & \na & \half & \na & $<$10\%\textsuperscript{h} & FP64 on INT8 \\
\midrule
& \textbf{Ours (CertiSum)} & \yes & \yes & \yes & \yes & \yes & \yes & \yes & \yes & 11.8--60.5\% & MXFP4 \\
\bottomrule
\end{tabular}
\label{tab:related}
\end{table*}

\begin{figure}[!t]
    \centering
    \includegraphics[width=\linewidth]{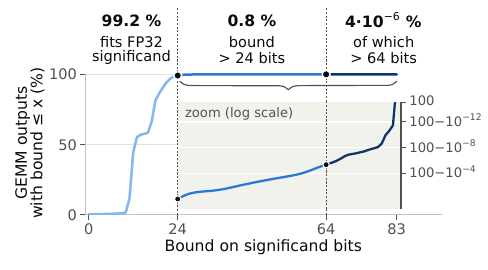}
    \caption{Upper bound on the significand bits needed for exact MXFP4 accumulation in any order, for GEMM outputs over 200 steps of training Qwen3-8B-Base on FineWeb-Edu. The bound uses the sum of absolute products; the shaded panel magnifies the tail (log scale).}
    \label{fig:gap6}
\end{figure}

Achieving these invariances at low overhead is difficult. Fixing the summation order does not suffice for hardware invariance, because vendors round low-precision GEMM intermediates differently~\cite{xie2025bit}, and the same order then yields different bits. Exact summation is invariant to both order and vendor, but it costs more than native FP32 accumulation, and whether that cost is necessary depends on run-time values. MXFP4 GEMMs typically accumulate in FP32~\cite{nvidia2026cutile,salykov2025matrixcores} and are exact only if every partial sum fits within the accumulator's 24-bit significand and exponent range~\cite{hayashi2026awe}. This depends on the operands' block scales, the reduction length, and the hardware's summation order. Exact methods that ignore run-time values must provision for the widest possible value spread~\cite{collange2015numerical}, paying this cost on every reduction even when native FP32 accumulation would have been exact. Methods that inspect values avoid this cost, but read both input matrices in an extra pass before every GEMM~\cite{schwarz2026guaranteed}.

Neither approach exploits the common case in which FP32’s significand is wide enough for exact accumulation (Figure~\ref{fig:gap6}). Our \textbf{key idea} is to certify this cheaply at run time, so that CertiSum keeps the output of the fast native MXFP4 GEMM and pays for exact computation only where certification fails or is not attempted. CertiSum evaluates increasingly expensive certificates, each a sufficient condition for exactness. The cheapest uses row statistics gathered in the same pass as quantization, so checking an output takes constant work independent of the reduction length. This works because MXFP4 elements have few significand bits under a shared per-block scale. On FineWeb-Edu, exact accumulation in any order needs at most 24 bits for 99.2\% of GEMM outputs. CertiSum skips the native GEMM for attention and nearly all weight-gradient GEMMs and runs them directly on exact kernels, which account for 46--68\% of training-step GEMM FLOPs (Appendix~\ref{sec:routing}).

We make the following key contributions.
\begin{itemize}
    \item We identify hardware invariance as a requirement for reproducible MXFP4 training, which existing methods either do not provide or provide at several times the runtime of native execution.
    \item We present CertiSum, a method that provides all three invariances for MXFP4 training with low overhead by certifying native MXFP4 GEMM results as exact at run time and computing the remaining outputs exactly.
    \item We show that CertiSum produces bitwise-identical Qwen3-8B-Base training across DP, TP, and CP layouts and across AMD MI350X and NVIDIA GB200. For Qwen3-8B-Base, Qwen3-14B-Base, and Llama-3.1-8B on FineWeb-Edu using AMD MI350X, it adds 11.8--60.5\% latency over its non-exact MXFP4 counterpart at the larger evaluated input shapes, where it remains faster than our BF16 and torchtitan baselines, and 15.4--77.3\% at the smaller ones.
\end{itemize}

\section{Background}
\label{sec:background}

\subsection{Microscaling Formats}
\label{sec:mx_formats}

MXFP4, the 4-bit member of the Microscaling (MX) format family, stores each value in 4.25 bits on average, $3.8\times$ fewer bits than BF16. This reduces arithmetic cost as well as the memory and bandwidth costs that limit LLM serving \cite{gholami2024ai}, while additional techniques mitigate model accuracy degradation~\cite{tseng2025training, rouhani2023microscaling, gao2025dpquant}. NVIDIA Blackwell and AMD MI350X GPUs support MXFP4 GEMMs natively \cite{nvidia2025nvfp4, amd2025cdna4}.

The OCP MX specification \cite{rouhani2023ocpmx} defines block-scaled formats in which $k$ low-bit elements $P_1, \dots, P_k$ share one scale $X$. Each element decodes to $v_i = X \cdot P_i$. MXFP4 uses $k = 32$, E2M1 elements, which take the values $0, \pm0.5, \pm1, \pm1.5, \pm2, \pm3, \pm4, \pm6$, and an E8M0 scale, which is a power of two. A block of wider-format values $V_1, \dots, V_k$ is quantized by choosing $X$ as a power of two from the block's largest magnitude and rounding each $V_i/X$ to the nearest E2M1 value.

Computing $AB^{\top}$, with MXFP4 matrices $A \in \mathbb{R}^{M \times K}$ and $B \in \mathbb{R}^{N \times K}$, requires $MN$ dot products of length $K$, each between a row of $A$ and a row of $B$.
To prevent excessive loss of accuracy \cite{wang2018training}, the terms of these dot products are accumulated in a wider format, and the specification recommends an FP32 result~\cite{rouhani2023ocpmx}.
The output is therefore not itself in MXFP4. Converting it to MXFP4 is a separate quantization step, needed only when the output is used in another MXFP4 operation \cite{rouhani2023microscaling}.

The specification fixes the encodings and the mathematical value of a block dot product, but leaves the accumulation precision, rounding behavior, and order of operations to the implementation~\cite{rouhani2023ocpmx, rouhani2023microscaling}. Two MX-compliant GEMM implementations can therefore return different bits for identical inputs.

\subsection{Reduction Order and Exactness}
\label{sec:reduction-order-and-exactness}

FP addition is non-associative, with $(a + b) + c \neq a + (b + c)$ in general. This is because floating-point addition rounds each result, $\operatorname{fl}(a+b) = (a+b)(1+\delta)$ with $|\delta| \le u$, where $u$ is the unit roundoff \cite{higham2002accuracy}. The rounding error $\delta$ is set by how far the exact sum $a+b$ lies from the nearest representable value, so different operands ($a$ and $b$) incur different rounding errors. Any change in the order of additions, which we call the \emph{reduction order}, may therefore change the operation's output bits.

The \emph{accumulation error} of a reduction $s = \sum_{i=1}^{n} x_i$ is given by $|\hat{s} - \operatorname{fl}(s)|$, where $\hat{s}$ is the computed sum and $\operatorname{fl}(s)$ is the exact sum $s$ rounded once to the output format. Measuring against $\operatorname{fl}(s)$ rather than $s$ excludes the unavoidable final rounding, so this error comes only from intermediate rounding.

The three invariances defined in \S\ref{sec:intro} impose different requirements on a reduction. Run-to-run invariance requires only a repeatable order, which floating-point atomic additions break. Fixing one reduction order regardless of the parallelism layout (\S\ref{sec:parallelism_layout}) achieves parallelization invariance; prior systems do so for changes in batch size or TP degree~\cite{he2025nondeterminism, zhang2025deterministic, jiang2026isoexec}. The accumulation error is then fixed but can be nonzero, so implementations that fix different orders can disagree. Hardware invariance requires every implementation to produce the same bits. \emph{Exact accumulation}, which sums without intermediate rounding in a sufficiently wide accumulator~\cite{hayashi2026awe}, achieves this for any order, because the correctly rounded sum is unique given identical terms and final rounding mode.

\subsection{Parallelism Layouts}
\label{sec:parallelism_layout}

Distributed ML training partitions the model and inputs across GPUs, and most of these partitions split reductions and change their order. We define a \emph{parallelism layout} as the set of factors, across GPUs and within each GPU, that determine the reduction order.

Across GPUs, each form of parallelism splits a reduction into per-device partial sums. Tensor parallelism (TP) splits GEMMs along their reduction dimension~\cite{shoeybi2019megatron}. Data parallelism (DP) splits the batch-dimension gradient sum across model replicas~\cite{cai2026flextrain}, and micro-batching splits each replica's share further. Context parallelism (CP) splits attention's key/value reduction across sequence chunks, in both the forward and backward passes~\cite{liu2024ringattention} (\S\ref{sec:blockwise_attention}). Changing the degree of any of these moves the split points and therefore changes the reduction order. A \emph{collective algorithm} then sets the order in which GPUs combine their partial sums. Libraries such as NCCL choose it from message size, GPU count, and network topology~\cite{hu2025demystifying}, and rebuilding gradient buckets after a change in GPU count can change the gradient aggregation order~\cite{li2023easyscale}.

Within a GPU, kernels split reductions across streaming multiprocessors (SMs) for throughput, as in split-K GEMMs~\cite{he2025nondeterminism, osama2023streamkworkcentricparalleldecomposition} and blockwise attention~\cite{dao2023flashdecoding} (\S\ref{sec:blockwise_attention}). Kernels choose this split based on the GEMM and micro-batch sizes.

\subsection{Blockwise Attention}
\label{sec:blockwise_attention}
Besides GEMMs, attention is the other dominant reduction in an LLM. Efficient implementations such as FlashAttention compute $\mathrm{softmax}(QK^{\top})V$ (omitting scaling by $1/\sqrt{d}$) blockwise, processing keys in blocks without materializing the full score matrix \cite{dao2022flashattention, liu2024ringattention}.
For each query row, attention tracks the running maximum $m$ of the scores seen so far. It forms $\tilde P = \exp(S - m)$, where $S=QK^{\top}$, and accumulates the unnormalized output $\tilde O =\tilde P V$ and the row sum $\ell = \sum_j \tilde P_j$, dividing by $\ell$ once at the end \cite{dao2023flashattention2}.
Whenever a new key block raises this running maximum, the partial $\tilde O$ and $\ell$ are rescaled by $\exp(m_{\text{old}} - m_{\text{new}})$ \cite{dao2023flashattention2}.

Softmax normalizes each query row over all keys, and the output sums the corresponding values, so every accumulated quantity is a reduction over the key/value dimension. 
Attention’s bitwise output therefore depends on where the key/value dimension is split, whether at kernel block boundaries, at CP-chunk boundaries, or at split-KV boundaries if used \cite{dao2023flashdecoding, he2025nondeterminism}. The order in which the kernel visits blocks also affects the bitwise output.

In the backward pass, $dQ$ sums over key blocks and $dK$, $dV$ sum over query blocks \cite{dao2023flashattention2}. Its reduction order therefore depends on both the query and key/value partitioning. Under CP, it also depends on the chunk layout and ring order \cite{liu2024ringattention}. 
Kernels that accumulate $dQ$ with atomic additions also make its order run-dependent \cite{dao2023flashattention2,he2025nondeterminism}.

\section{CertiSum}
\label{sec:CertiSum}

\begin{figure}[!t]
  \centering
  \includegraphics[width=\columnwidth]{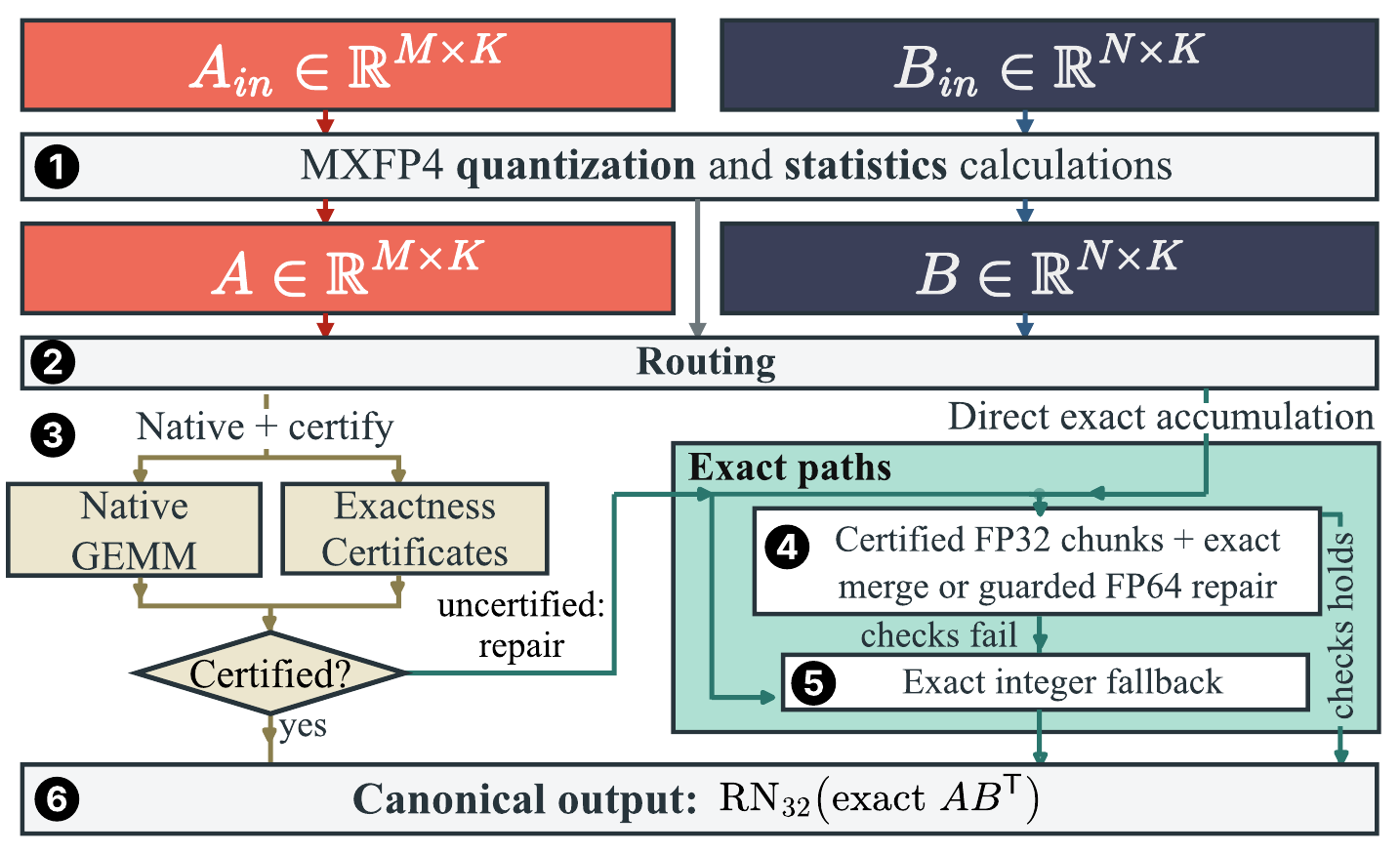}
  \caption{CertiSum GEMM execution; steps \dingOne--\dingSix\ described in \S\ref{sec:design_details}.}
  \label{fig:CertiSum-implementation}
\end{figure}

We introduce CertiSum, a method for exact accumulation of MXFP4 values and their products without prescribing a reduction order. We first explain certification and exact computation, then establish the conditions under which exact merging preserves results across execution layouts, using GEMM and attention as applications. Figure~\ref{fig:CertiSum-implementation} gives an overview of how CertiSum executes a GEMM.

\textbf{Design Challenge~\dingone: cheap certification.} Exactness depends on the current operand values and block scales. Checking individual product magnitudes would require another reduction for every output. We need reusable input statistics that can certify many outputs without recomputing their dot products.

\textbf{Design Challenge~\dingtwo: soundness and layout invariance.} A certificate must cover every reduction order and remain sound when its own evaluation uses rounded arithmetic. Across devices, exact local sums are insufficient if repartitioning changes MX blocks or if partial sums are rounded before merging. The mechanism must preserve both the reduction terms and their exact partial sums.

\subsection{A Sufficient Condition for Native Exactness}
\label{sec:CertiSum-condition}

We write GEMM as $AB^\top$, where $A\in\mathbb{R}^{M\times K}$ and $B\in\mathbb{R}^{N\times K}$ contain the decoded MXFP4 inputs. For output position $(i,j)$, let $x=A_{i,:}$ and $y=B_{j,:}$ be the corresponding input rows, so that $x_k=A_{ik}$ and $y_k=B_{jk}$. Their exact dot product is $s=\sum_{k=1}^{K}x_ky_k$. However an output is computed, CertiSum returns $c_{32}=\operatorname{RN}_{32}(s)$, or $c_{\mathrm{bf16}}=\operatorname{RN}_{\mathrm{bf16}}(c_{32})$ for BF16 output, where $\operatorname{RN}_f$ denotes round-to-nearest-even in format $f$ as defined by IEEE~754~\cite{ieee2019754}. These targets refer to the quantized operands; exact zero is returned as $+0$. We call a native MXFP4 kernel with FP32 accumulation \emph{natively exact} at $(i,j)$ if its output equals this target for every order and grouping of products and additions it may use.

To bound the precision required by every reduction order, express each row in a common power-of-two unit. For a nonzero row $x$, let $\lambda_x$ be half the smallest MX scale among blocks containing nonzero values, and define $\lambda_y$ analogously. Then $x_k=\lambda_x a_k$ and $y_k=\lambda_y b_k$ for integers $a_k,b_k$. Zero rows return zero directly. With $\lambda=\lambda_x\lambda_y$, the dot product is $s=\lambda\sum_k a_kb_k$. We seek a conservative upper bound $\overline{T}$ satisfying
\begin{equation} T \equiv \sum_k |a_kb_k| \;\leq\; \overline{T} \;<\; 2^{24}. \label{eq:CertiSum-certificate} \end{equation}

This condition bounds every partial sum, including those formed before cancellation. For any subset $I$, $|\sum_{k\in I}a_kb_k|\leq T$. Each partial sum therefore needs at most 24 significant bits in the common unit. Range checks on the unit $\lambda$, which we call \emph{exponent guards}, keep nonzero products and partial sums normal and finite in FP32. We assume native kernels that initialize each dot-product accumulator to zero, return exactly representable FP32 products and sums unchanged, and implement the prescribed final conversion without intermediate BF16 rounding. Under these conditions, every product and addition is exact regardless of the reduction tree; this is the error-free condition of Ozaki-type schemes~\cite{ozaki2012error,hayashi2026awe}, here checked at run time. Appendix~\ref{sec:appendix-native} gives the exponent guards and proves native exactness under these conditions.

\subsection{Evaluating Certificates from Reusable Statistics}
\label{sec:CertiSum-statistics}

\begin{figure}[!t]
    \centering
    \includegraphics[width=\linewidth]{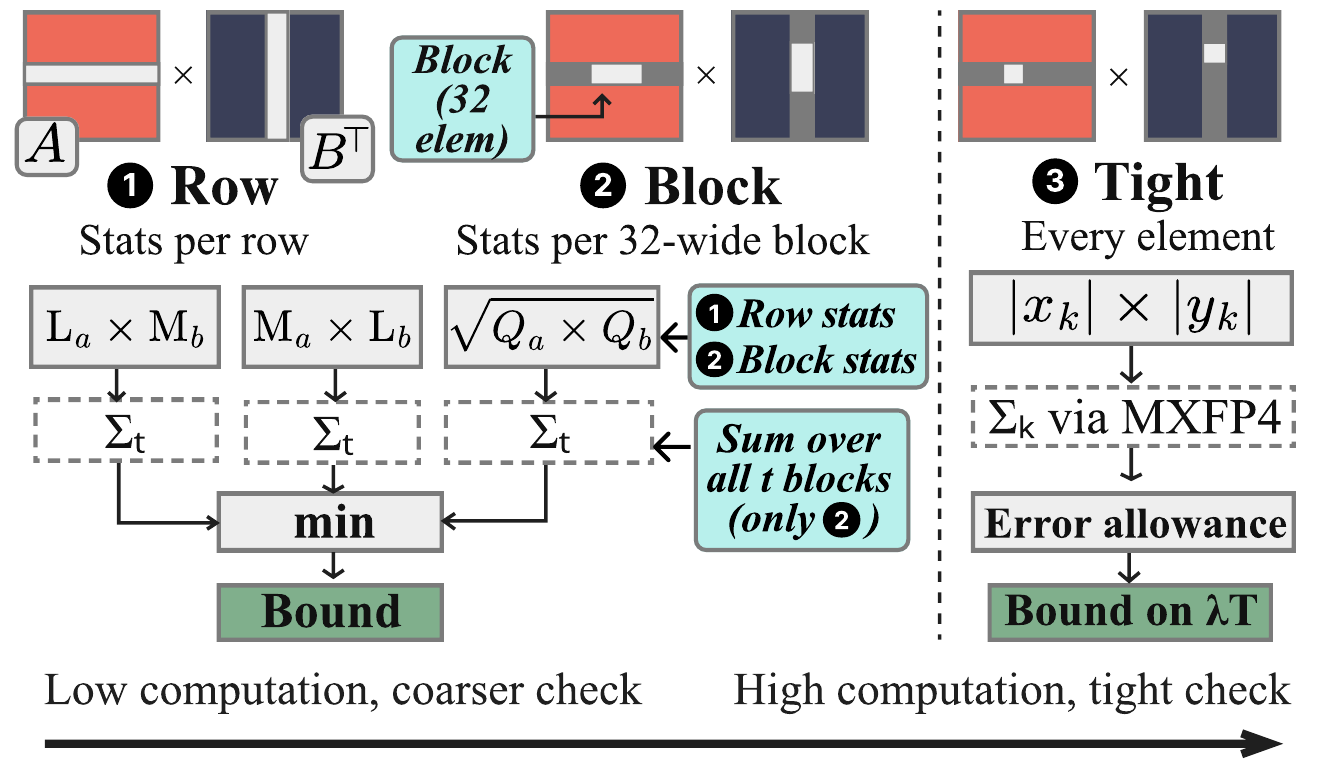}
    \caption{Certificates from row (\dingOne), block (\dingTwo), and absolute-product (\dingThree) statistics, in increasing cost and tightness.}
    \label{fig:certificates}
\end{figure}

To check Equation~\eqref{eq:CertiSum-certificate} at a cost independent of the reduction length $K$, instead of computing the $K$ products $a_kb_k$ of every output, CertiSum collects row statistics during quantization. For either integer row $v\in\{a,b\}$, these are $L_v=\sum_k|v_k|$, $M_v=\max_k|v_k|$, and $Q_v=\sum_k v_k^2$. These statistics summarize each entire input row and give the following \emph{row bound} (Figure~\ref{fig:certificates}, \dingOne) for the selected output: \begin{equation} B_{\mathrm{row}}=\min\!\left\{L_aM_b,\; M_aL_b,\;\sqrt{Q_aQ_b}\right\}\geq T. \label{eq:CertiSum-row-bound} \end{equation}
The first term bounds every $|b_k|$ by $M_b$; the second exchanges the rows; the third follows from Cauchy--Schwarz. Componentwise maxima of the statistics over groups of rows can certify several outputs together.

When the row bound fails to certify an output, CertiSum tries two tighter but more costly bounds (Figure~\ref{fig:certificates}, \dingTwo\ and \dingThree). \emph{Block bounds} restrict each of the three expressions in Equation~\eqref{eq:CertiSum-row-bound} to 32-element blocks along $K$. CertiSum sums each expression over blocks, then takes the minimum of the three sums (Appendix~\ref{sec:impl-mechanism}). Across all outputs, each sum is a matrix multiplication with reduction length $K/32$. The \emph{tight bound} evaluates $\lambda T=\sum_k|x_ky_k|$ using a native MXFP4 GEMM on sign-cleared operands with unchanged scales. If all three bounds fail, CertiSum repairs the output (\S\ref{sec:CertiSum-merging}). For shapes that rarely certify, CertiSum skips the native GEMM and computes all outputs exactly (\S\ref{sec:impl-pipeline}).

The row, block, and tight bounds are themselves computed with rounded arithmetic, so CertiSum evaluates each one conservatively, accounting for rounding of the statistics, products, accumulation, and output conversion. In particular, the native MXFP4 GEMM that evaluates the tight bound may itself round, so its verified error allowance must still produce an upper bound $\overline{T}\geq T$. Appendix~\ref{sec:appendix-certificates} proves the bound hierarchy and derives sufficient rounding allowances under explicit kernel assumptions.

\subsection{Repair and Exact Merging}
\label{sec:CertiSum-merging}

When an output fails certification, where its exactness is unsure, CertiSum recomputes it exactly from the same quantized operands. It has three ways to do so. The first splits the reduction into chunks short enough to pass the native certificate, sums each chunk with the native kernel, and merges the chunk sums exactly, as described below. The second accumulates the products in FP64, as in emulated GEMMs~\cite{ozaki2012error,schwarz2026guaranteed}, provided a \emph{capacity check} shows that no partial sum can round. The third, the \emph{integer path}, accumulates in multi-word integers, a superaccumulator~\cite{kulisch2013computer,iakymchuk2016exblas}, and is the fallback when the other two fail their checks (Appendix~\ref{sec:appendix-repair}). Direct exact execution uses the same three methods.

Exact merging combines partial sums from different chunks or devices without rounding, so the result does not depend on how the reduction was split. Formally, let disjoint index sets $J_r$ partition a dot product, and let $\lambda Z_r$ be the exact partial sum of partition $r$. After aligning units by exact power-of-two shifts, CertiSum computes
\begin{equation} Z_r=\sum_{k\in J_r}a_kb_k,\qquad c_{32}=\operatorname{RN}_{32}\!\left(\lambda\sum_r Z_r\right). \label{eq:CertiSum-merge} \end{equation}
A capacity check verifies that $\sum_r |Z_r|$ stays below $2^{31}$ for a signed int32 accumulator or $2^{53}$ for an FP64 accumulator. Every partial sum of the $Z_r$ then fits in the accumulator, so these additions do not round. Integer addition is associative and commutative, so the result is the same for every partition size, merge order, and reduction tree, the argument behind long-accumulator and reproducible summation~\cite{kulisch2013computer,collange2015numerical,ahrens2020algorithms}. Partitions therefore keep their exact partial values until the merge is complete. Rounding a partial sum to the output format first could lose bits that the final result needs. Appendix~\ref{sec:appendix-repair} gives sufficient conditions for exact repair and proves partition invariance.

\subsection{Preserving Reductions Across Layouts}
\label{sec:CertiSum-layouts}

Exact merging requires every layout to sum the same products $a_kb_k$. These products would otherwise depend on the parallelism layout: the MX block an element belongs to sets its scale, and blocks formed within a device's shard change with the shard. CertiSum therefore assigns elements to MX blocks using their positions in the full tensor, not their positions within each device's shard, including the MX blocks along the token dimension used to quantize weight-gradient operands in MX training recipes~\cite{rouhani2023microscaling}. Tensor-parallel shards and data-parallel or microbatch contributions to a weight gradient are then partial sums over disjoint parts of one reduction. CertiSum keeps each exact and adds them without rounding (\S\ref{sec:CertiSum-merging}), scaling every contribution by the same global loss normalization.

Attention (\S\ref{sec:blockwise_attention}) also merges partial sums across partitions, and it ends with a division by a sum, so exact merging must cover both the numerator and the denominator. Query--key scores use the certified or exact dot products above. Unnormalized attention weights and values are quantized in MX blocks fixed by global key and token positions, so splitting keys across devices or kernel blocks yields the same quantized weights (Appendix~\ref{sec:impl-mechanism}). To bound the range of retained attention weights, CertiSum prunes key blocks whose scale falls too far below a fixed per-row reference. The reference does not depend on the key partition, so neither does pruning; FoldAttention~\cite{achanta2026foldattention} fixes a softmax reference for the same reason.

For query $i$ and key $j$, let $w_{ij}\geq0$ and $\mathbf{v}_{ij}$ denote the resulting quantized unnormalized attention weight and value vector. Masked or pruned keys have $w_{ij}=0$. CertiSum computes the attention-weight sum $\ell_i$ and the $w$-weighted sum of value vectors $\mathbf{n}_i$ exactly. It obtains the attention output vector $\mathbf{o}_i$ by rounding $\mathbf{n}_i/\ell_i$ componentwise to FP32:
\begin{equation} \begin{aligned} \mathbf{n}_i&=\sum_j w_{ij}\mathbf{v}_{ij},\qquad \ell_i=\sum_j w_{ij}>0,\\ \mathbf{o}_i&=\operatorname{RN}_{32}\!\left(\mathbf{n}_i/\ell_i\right). \end{aligned} \label{eq:CertiSum-attention} \end{equation}
Key partitions merge their numerator and denominator sums exactly, then round the exact quotient componentwise, the split-key decomposition of blockwise attention~\cite{dao2023flashdecoding,liu2024ringattention} with exact instead of rounded merging. Backward contributions use the same exact merge rule.

Exact GEMMs and attention alone do not make a whole training step invariant. Every layout must also see the same batch, start from the same parameters and optimizer state, and use the same random choices, masks, loss normalization, and rounding points. Deterministic training frameworks already fix the batch order, the state, and the random choices~\cite{aithal2026megatrondeterminism,li2023easyscale}. Operators outside these reductions must implement identical numerical maps across supported hardware. CertiSum commits an update only after every required output satisfies its numerical contract and every failed range or capacity check on any GPU has been resolved by repair or recomputation. Under the conditions above, any two accepted executions of the same logical step produce identical parameter and optimizer-state bits. Appendix~\ref{sec:appendix-invariance} proves the attention and training-step guarantees.
\section{Implementation Details} 
\label{sec:design_details}
\label{sec:impl-pipeline}

For each GEMM, CertiSum chooses between native execution with certification and direct exact computation. Both routes return the same target (\S\ref{sec:CertiSum-condition}); only the cost differs. This section walks through the numbered steps of Figure~\ref{fig:CertiSum-implementation}.

\textbf{Operand preparation \dingOne.} To reuse the operand reads needed for quantization, activation quantization produces the 4-bit MXFP4 elements, block scales, and reusable certificate statistics together. Weights change only at optimizer updates, so CertiSum caches their statistics between updates. The decoded quantized operands are $A$ and $B$; $A_{\mathrm{in}}$ and $B_{\mathrm{in}}$ denote their inputs before quantization.

\textbf{Route selection \dingTwo.} Requiring a full native GEMM before every exact computation would add work even when direct exact execution is preferable. Tensor shapes and previous certificate outcomes guide the choice between native execution with certification (\dingThree) and direct entry into the exact paths (\dingFour\ or \dingFive). This choice determines which work is attempted; acceptance still requires checks on the current operands.

\textbf{Native execution and certification \dingThree.} Outputs left uncertified by the selected checks of \S\ref{sec:CertiSum-statistics} enter the exact paths (\dingFour\ or \dingFive) for repair; failure to certify does not establish that a value is incorrect. To avoid a CPU routing decision before repair, CertiSum supports in-stream repair, whose kernels read flags stored on the GPU and return immediately when no repair is needed.

\textbf{Exact computation \dingFour\,--\,\dingFive.} These paths implement the repair methods of \S\ref{sec:CertiSum-merging}, with chunked native accumulation and FP64 accumulation in \dingFour\ and the integer path in \dingFive, which routing can also select directly. The certified or exact result supplies the GEMM output (\dingSix).

\section{Evaluation}
\label{sec:evaluation}

\subsection{Experimental Setup}
\label{sec:experimental-setup}

\textbf{Hardware platforms.}
We evaluate CertiSum on the two hardware platforms listed in Table~\ref{tab:hardware}. Both support MXFP4 GEMMs natively in their matrix units.

\begin{table}[t]
\centering
\small
\caption{Hardware platforms.}
\vspace{4pt}
\setlength{\tabcolsep}{2pt}
\begin{tabular}{@{}lccl@{}}
\toprule
Platform & GPUs & HBM3E/GPU & Used in \\
\midrule
\makecell[l]{AMD Instinct MI350X\\\cite{amd2025cdna4}} & 1, 2, 4 & 288\,GB & all \\
\addlinespace[2pt]
\makecell[l]{NVIDIA GB200\\\cite{nvidia2025blackwell}} & 1, 2, 4 & 186\,GB & \S\ref{sec:cross-platform} \\
\bottomrule
\end{tabular}
\label{tab:hardware}
\end{table}

\textbf{Datasets.}
We train on FineWeb-Edu~\cite{penedo2024fineweb}, a corpus of educational web text, and The Stack v3~\cite{lozhkov2026stack-v3,lozhkov2024starcoder}, a source-code corpus. FineWeb-Edu inputs come from the \texttt{sample-10BT} subset.

\textbf{Models.}
We train the base models Qwen3-8B-Base, Qwen3-14B-Base~\cite{yang2025qwen3}, and Llama-3.1-8B~\cite{grattafiori2024llama}. The inference experiments in \S\ref{sec:inference} use post-trained Qwen3-8B~\cite{yang2025qwen3}.

\textbf{Parallelism layouts.}
 A suffix gives the number of GPUs. For example, TP2 denotes two-way tensor parallelism.

\textbf{Input shapes.}
We denote input shapes as $B\times S$, where $B$ is the number of sequences and $S$ is the tokens per sequence. Thus, $4\times4096$ denotes four sequences of 4,096 tokens each. Shapes are global unless explicitly marked per GPU, as in the DP4 rows of Table~\ref{tab:mi350x-training}.

\textbf{Implementations.}
Table~\ref{tab:arms} lists the compared implementations and the labels that \S\ref{sec:performance} uses for them.

\begin{table}[htb]
\centering
\small
\caption{Compared implementations.}
\vspace{4pt}
\setlength{\tabcolsep}{4pt}
\begin{tabular}{@{}l p{0.82\columnwidth}@{}}
\toprule
Label & Implementation \\
\midrule
$E$ & CertiSum; its native MXFP4 GEMM path uses AITER~\cite{amd2025aiter} on MI350X. At the smaller input shapes, the faster measured result of ordinary (eager) execution and a static-route variant that replays recorded GPU operations. \\
\addlinespace[2pt]
$T$ & MXFP4 twin: CertiSum's structure with non-exact MXFP4 GEMMs and non-exact versions of our attention kernels. \\
\addlinespace[2pt]
$F$ & BF16: the faster measured result of our BF16 implementation and torchtitan~\cite{liang2025torchtitan}. \\
\addlinespace[2pt]
$D$ & $F$ in PyTorch deterministic mode~\cite{pytorch2026deterministic}; deterministic torchtitan uses SDPA attention. \\
\addlinespace[2pt]
Vendor & Native MXFP4 GEMMs through hipBLASLt~\cite{amd2026hipblaslt} with our pinned Tensile library on MI350X and cuBLASLt~\cite{nvidia2026cublaslt} on GB200, with BF16 FlashAttention on dequantized MXFP4 inputs (AITER on MI350X, FlashAttention-4~\cite{zadouri2026flashattention} on GB200). \\
\bottomrule
\end{tabular}
\label{tab:arms}
\end{table}

Appendix~\ref{sec:appendix-experimental} gives further experimental details.

\subsection{Cross-Platform Invariance}
\label{sec:cross-platform}

To test whether CertiSum preserves a training trajectory across GPU vendors and GPU counts, we train Qwen3-8B-Base on FineWeb-Edu and The Stack v3 for 200 AdamW steps~\cite{loshchilov2017decoupled}. Each step uses $1\times4096$. We run on AMD MI350X and NVIDIA GB200 using one GPU and TP4 on each platform. In this experiment, TP4 divides each GEMM's output rows and the attention heads among four GPUs and splits no reduction across GPUs.

We compare CertiSum with the vendor baseline of Table~\ref{tab:arms}. All runs start from identical model and optimizer states and consume identical token batches in the same order.

\begin{figure}[htb]
  \centering
  \includegraphics[width=\columnwidth]{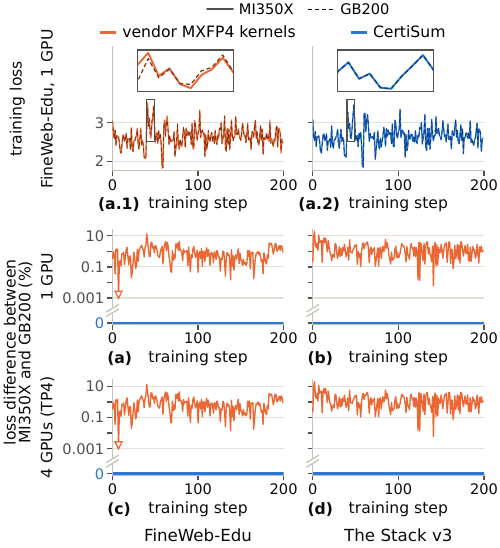}
  \caption{\textbf{Cross-Platform Invariance (MI350X vs. GB200).} (a.1), (a.2): FineWeb-Edu training loss on one GPU for the vendor baseline (Table~\ref{tab:arms}) and CertiSum. (a)--(d): relative cross-vendor loss difference on a log scale; exact zeros below the axis break; triangles mark differences below $0.001\%$.}
  \label{fig:cross-platform-invariance}
\end{figure}

Figure~\ref{fig:cross-platform-invariance} shows training losses above and cross-vendor relative loss differences below. For each method, the latter is $100|L_{\mathrm{GB200}}-L_{\mathrm{MI350X}}|/L_{\mathrm{MI350X}}$, where $L$ is the loss at the corresponding training step. All curves are unsmoothed.

\begin{figure}[!t]
  \centering
  \includegraphics[width=\columnwidth]{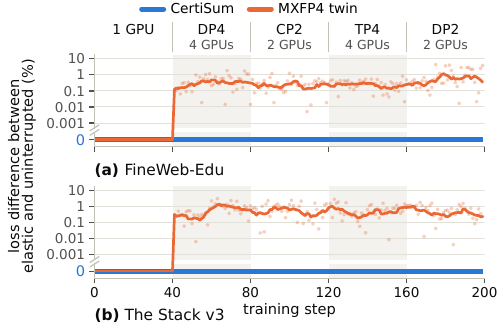}
  \caption{\textbf{Elastic Training.} Relative loss difference from the uninterrupted one-GPU reference on (a) FineWeb-Edu and (b) The Stack v3. Dots are steps; lines are a centered nine-step geometric mean of positive values; zeros are below the axis break.}
  \label{fig:elastic-training}
\end{figure}

As shown in Figure~\ref{fig:cross-platform-invariance}, CertiSum produces bitwise-identical losses on AMD and NVIDIA at all 200 steps, with one GPU and TP4, on both datasets. Per-step tensor fingerprints and final parameter and optimizer-state hashes match as well, so the whole training state is invariant, not just the loss.

The baseline's AMD and NVIDIA losses, in contrast, differ from step 0 onward. Median differences are 0.75\% and 1.19\% on FineWeb-Edu and The Stack v3, respectively, reaching 13.34\% and 20.10\%.

Under shared numerical definitions, CertiSum therefore trains bitwise identically on both vendors and both GPU counts, satisfying the hardware invariance defined in \S\ref{sec:intro}.

\begin{figure*}[!t]
  \centering
  \includegraphics{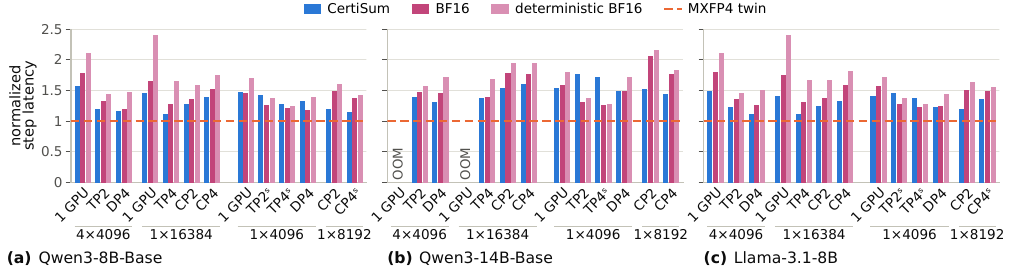}
  \caption{Training-step latency on FineWeb-Edu using AMD MI350X, normalized to the MXFP4 twin (dashed line); lower is better. BF16 bars use the faster measured implementation per mode (Table~\ref{tab:arms}). Shapes are sequences $\times$ tokens, global except DP4 (per GPU). Superscript $s$: static-route CertiSum; OOM: all implementations ran out of memory.}
  \label{fig:performance}
\end{figure*}

\subsection{Elastic Training}
\label{sec:elastic}

To test whether CertiSum preserves the training trajectory across changes in parallelism layout, we change the GPU count and layout during a run. We reuse the training setup of \S\ref{sec:cross-platform} on AMD MI350X GPUs, with a global batch of $4\times4096$. Each elastic run resumes from checkpoints between 40-step segments, following $1\,\mathrm{GPU}\rightarrow\mathrm{DP4}\rightarrow\mathrm{CP2}\rightarrow\mathrm{TP4}\rightarrow\mathrm{DP2}$. Controls checkpoint and resume at the same boundaries while remaining on one GPU.

Figure~\ref{fig:elastic-training} compares CertiSum and its non-exact MXFP4 twin with their respective uninterrupted one-GPU references. It plots relative loss difference against training step. CertiSum matches its reference's loss bits at all 200 steps on both datasets. Its checkpointed parameters and optimizer states match the one-GPU control bit for bit. The non-exact MXFP4 twin's losses, in contrast, diverge at step 41, after the first layout change, and remain different thereafter, reaching 4.2\% on FineWeb-Edu and 3.0\% on The Stack v3. Both methods' one-GPU controls reproduce the uninterrupted loss bits, so checkpointing alone is not the cause.

Resizing across one GPU, DP, CP, and TP leaves CertiSum's trajectory untouched, so parallelization invariance holds.

\begin{figure}[!t]
  \centering
  \includegraphics[width=\columnwidth]{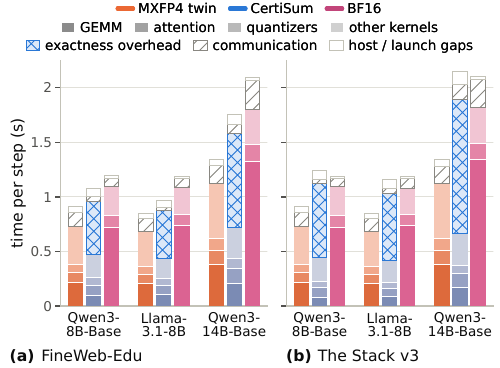}
  \caption{Traced step time on four AMD MI350X GPUs (DP4, $4\times4096$ per GPU) on (a) FineWeb-Edu and (b) The Stack v3; bars are the MXFP4 twin, CertiSum, and BF16. Each bar is the slowest GPU, averaged over steps 3--28 of one run. Compute segments are scaled from GPU-averaged kernel sums; crosshatching marks certification, repair, and exact GEMMs; communication excludes overlap; unfilled segments are host/launch gaps and logging.}
  \label{fig:mi350x-runtime-breakdown}
\end{figure}

\subsection{Inference Invariance Across Batching and Layouts}
\label{sec:inference}

To test whether batching and GPU layout change the tokens generated during inference, we use 32 general-knowledge questions. These questions were generated with Claude Code and are listed verbatim in Appendix~\ref{sec:appendix-questions}. We greedily generate 64 tokens per question using post-trained Qwen3-8B on AMD MI350X GPUs. We run each question alone on one GPU, in a batch of four on one GPU, and in the same batch with TP4. Table~\ref{tab:inference-invariance} reports how many generated sequences match each method's own single-question, one-GPU reference.

\begin{table}[htb]
\centering
\small
\caption{Generated sequences (out of 32) identical to each method's batch-1, one-GPU reference.}
\vspace{4pt}
\setlength{\tabcolsep}{4pt}
\begin{tabular}{@{}lcc@{}}
\toprule
Method & Batch 4, 1 GPU & Batch 4, TP4 \\
\midrule
CertiSum & \textbf{32}/32 & \textbf{32}/32 \\
MXFP4 twin & 31/32 & 0/32 \\
Deterministic BF16 & 29/32 & 25/32 \\
\bottomrule
\end{tabular}
\label{tab:inference-invariance}
\end{table}

In both batched configurations, CertiSum generates the same 32 sequences as its one-GPU reference, and all 2,048 FP32 log-probability vectors match bit for bit. All three methods reproduce their batch-1 tokens and log-probabilities on repetition, yet deterministic BF16 changes 3 and 7 sequences and the twin 1 and 32, respectively.

Batch composition and TP degree do not change CertiSum's generated tokens or log-probabilities, so parallelization invariance extends to inference.

\subsection{Overhead}
\label{sec:performance}

To measure the latency overhead of CertiSum, we time training steps of the three models using one to four AMD MI350X GPUs on FineWeb-Edu, comparing implementations $E$, $T$, $F$, and $D$ of Table~\ref{tab:arms}. We report the median over runs of the mean latency of steps 3--29, taking the slowest GPU at each step; Appendix~\ref{sec:appendix-performance} gives the run counts and exclusion rule.

Figure~\ref{fig:performance} plots the results; Table~\ref{tab:mi350x-training} in Appendix~\ref{sec:appendix-performance} reports $E$ and $D$ in ms and the ratios $E/T$, $E/F$, and $E/D$ for 37 timed model, layout, and input-shape configurations. The main shapes are the larger ones ($4\times 4096$ and $1\times 16384$). CertiSum adds 11.8--60.5\% latency over its MXFP4 twin at the main shapes and 15.4--77.3\% at the smaller shapes.

At the main shapes, CertiSum has lower measured median latency than both ordinary BF16 references. Its latency is 0.81--1.00$\times$ that of the faster reference, corresponding to a speedup of up to $1.24\times$. Qwen3-14B-Base TP4 nearly matches this reference (992.3 versus 994.6\,ms). Against the faster deterministic BF16 reference, its latency is 0.59--0.89$\times$, corresponding to a speedup of up to $1.70\times$.

For smaller inputs, we observe that the advantage over BF16 can reverse. This is likely because certification and routing add kernel launches and host work with a fixed cost per GEMM. As the GEMMs shrink, this cost becomes a larger share of the step. All six smaller TP configurations take 1.05--1.36$\times$ the faster ordinary BF16 reference's latency. In deterministic mode, the range is 1.03--1.34$\times$. All six smaller CP configurations retain lower latency, at 0.74--0.92$\times$ and 0.70--0.88$\times$ the respective references. Qwen3-8B-Base on one GPU and with DP4 also takes longer than the faster ordinary BF16 reference at the smaller shapes, at 1.01$\times$ and 1.13$\times$, but remains faster than deterministic BF16.

On FineWeb-Edu using AMD MI350X, CertiSum is therefore faster than the measured BF16 references at the main input shapes that fit in memory but not at the smaller TP shapes.

\subsection{Runtime Breakdown}

To identify how the training workload affects CertiSum's runtime, we profile Qwen3-8B-Base, Llama-3.1-8B, and Qwen3-14B-Base on FineWeb-Edu and The Stack v3. For each model, we hold the GPU layout, batch size, and sequence length fixed across datasets. We compare CertiSum with its non-exact MXFP4 twin and our BF16 baseline.

Figure~\ref{fig:mi350x-runtime-breakdown} shows traced time per step in seconds by runtime component for each model. These durations include profiling overhead; \S\ref{sec:performance} reports untraced performance.

Moving from FineWeb-Edu to The Stack v3 increases CertiSum's traced step time by 16--22\%, while the twin and BF16 change by less than 0.5\%. Within CertiSum, certification, repair, and exact GEMM computation show the largest component increase for every model. In the underlying traces, this group accounts for 83--88\% of the increase in summed compute-kernel duration. This percentage describes kernel work before accounting for overlap.

\section{Discussion}
\label{sec:discussion}

\textbf{Other low-precision formats.} We rely on properties of block-scaled formats rather than of MXFP4 specifically: quantized elements are integers in a common power-of-two unit, a bound on their products certifies native accumulation, and exact paths merge partial sums without rounding. MXFP6~\cite{rouhani2023ocpmx} keeps the power-of-two block scales of MXFP4 and therefore this integer representation. NVFP4~\cite{nvidia2025nvfp4} preserves it as well: an E4M3 block scale is a small integer times a power of two, and the FP32 tensor scale, if computed from the whole tensor, is a common factor that can be applied once to the exact sum. Both would need format-specific bounds and kernels. MXFP8 fits the same design, but its wider elements make native certification succeed less often and shift more work to the exact paths.

\textbf{Integration into existing frameworks.} CertiSum replaces a framework's GEMM, attention, and gradient-collective operators, their autograd definitions, and the end-of-step update commit (\S\ref{sec:CertiSum-layouts}); frameworks with custom operators and autograd functions, such as PyTorch, support these replacements, and their remaining operators must meet the conditions stated there. RL systems that train and sample in different frameworks must install it in both, and audits must record that it was enabled.

\section{Related Work}
\label{sec:related_work}

Table~\ref{tab:related} compares CertiSum with prior systems; this section extends the discussion of \S\ref{sec:intro}.
\textbf{Order-fixing systems}~\cite{cai2026flextrain, bai2026finish, gond2026llm42} produce identical results only if the order of reductions is fixed, so none makes training invariant to a change of the TP, DP, or CP degree. Verde~\cite{arun2025verde} fixes reduction orders across hardware but assumes consistent rounding, which low-precision FP lacks~\cite{xie2025bit}, and does not cover inter-GPU reductions.
\textbf{Order-free summation} avoids fixed ordering by using wide accumulation, in software~\cite{collange2015numerical, iakymchuk2016exblas} or with hardware support~\cite{lutz2017high}.
\textbf{Cross-hardware reproduction} systems reproduce a given GPU workload deterministically on other hardware, such as CPUs~\cite{badash2026hawkeye, donaghy2026open1bfullyauditabletraining, xie2025repdl}, but do not provide parallelization invariance on their own.
\textbf{Exact GEMMs} have precedent in GEMM emulation~\cite{ozaki2012error, schwarz2026guaranteed, hayashi2026awe}, which splits operands into low-precision components whose products fast matrix routines compute without rounding.
AWE~\cite{hayashi2026awe} fixes its encoding offline and must therefore provision for worst-case input magnitudes, whereas CertiSum certifies exactness from the run-time operands (\S\ref{sec:CertiSum}).
ADP~\cite{schwarz2026guaranteed} also inspects run-time inputs, but chooses its slice count, scaling, and FP64 fallback per call from that call's operands, so partitioning a reduction differently can change the output bits.

\section{Conclusion}

In this work, we present CertiSum for bitwise-invariant MXFP4 training across runs, parallelism layouts, and GPU vendors. It certifies when native FP32 accumulation of MXFP4 products is exact for every summation order and uses exact kernels where needed. Fixed quantization blocks and exact merging preserve reduction results across layouts. CertiSum preserves training states bit for bit across AMD and NVIDIA GPUs and through DP, TP, and CP changes, while often remaining faster than BF16 training. CertiSum thus lets low-precision training move across GPU counts, parallelism layouts, and vendors without changing its results, enabling exact reproduction for debugging, matching training and inference engines in RL, and training audits on different accelerators.

\bibliographystyle{mlsys2025}
\bibliography{references/refs}

\section*{AI Usage Disclosure}

The authors used ChatGPT and Claude through their web interfaces, and Claude Code through both Claude Desktop and the command-line interface. Claude Code assisted with implementation and kernel optimization, and the authors reviewed the generated code. AI tools assisted with the implementation of plotting scripts for figures in the introduction and evaluation. The authors checked these scripts for correctness. The authors conducted the literature search, then used Claude and ChatGPT to check for omissions. The authors wrote the text in the paper and used AI tools to edit it. The authors developed research ideas and experimental design without AI assistance. The authors used Claude Code to generate the 32 general-knowledge questions used for the inference experiments. AI use for data generation was limited to these questions, which are listed verbatim in the appendix.
\clearpage
\appendix
\section{Mechanism Details}
\label{sec:impl-mechanism}

This section details two mechanisms that \S\ref{sec:CertiSum} summarizes.

\textbf{Block bounds.} Block bounds restrict the row statistics of \S\ref{sec:CertiSum-statistics} to each 32-element block $t$ along $K$, giving $L_{v,t}$, $M_{v,t}$, and $Q_{v,t}$ for $v\in\{a,b\}$. Each term of Equation~\eqref{eq:CertiSum-row-bound}, summed over blocks, bounds $T$; for example, $T\leq\sum_t L_{a,t}M_{b,t}$.

\textbf{Attention quantization.} Unnormalized attention weights are quantized in fixed global 32-key blocks, each scaled from its valid scores. To bound how small retained weights can be, CertiSum prunes blocks relative to a fixed row reference, which is the larger scale exponent of the first and query-containing global blocks. This reference fixes rescaling and pruning across key partitions. Value quantization follows global token positions. Backward quantization uses fixed $32\times32$ query--key tiles and shares the quantized score gradient between query and key gradients.

\section{Exactness and Invariance Guarantees}
\label{sec:appendix}

We prove the guarantees used in \S\ref{sec:CertiSum}, with its notation $x_k=\lambda_x a_k$, $y_k=\lambda_y b_k$, $\lambda=\lambda_x\lambda_y$, $s=\lambda\sum_k a_kb_k$, and $T=\sum_k|a_kb_k|$. Inputs have non-NaN MX scales and finite FP32-representable decoded values; zero rows return $+0$. All statements concern the quantized operands. Acceptance requires the stated kernel, range, and capacity checks.

\subsection{Native Exactness}
\label{sec:appendix-native}

E2M1 elements are integer multiples of $\tfrac12$ with magnitude at most six in the OCP MX specification~\cite{rouhani2023ocpmx}. Since MX scales are powers of two, choosing half the smallest nonzero-block scale as each row's unit makes $a_k,b_k$ integers. Thus every product and partial sum belongs to $\lambda\mathbb{Z}$. The sufficient condition below is the error-free argument underlying the Ozaki scheme~\cite{ozaki2012error}, which fixes slice widths in advance so that every slice product is exact; CertiSum instead checks a data-dependent bound for each output at run time.

\textbf{Proposition 1 (Native exactness).} Suppose the native kernel starts at zero, returns the exact result of every product and of every addition it performs, whether a binary addition or a fused addition of several aligned terms, whenever that result is representable in FP32, and performs no intermediate BF16 conversion. Sufficient conditions for exact accumulation under every reduction tree are
\begin{equation}
T\leq\overline{T}<2^{24},\qquad 2^{-126}\leq\lambda\leq2^{103}.
\label{eq:appendix-native-guards}
\end{equation}
The kernel must implement the prescribed final conversion and return $+0$ for exact zero.

\emph{Proof.} Each tree node sums a subset $I$ of the products and has exact value $\lambda z_I$, where $z_I=\sum_{k\in I}a_kb_k\in\mathbb{Z}$ and $|z_I|\leq T<2^{24}$. It therefore has at most 24 significant binary digits. If nonzero, its magnitude is at least $2^{-126}$ and at most $(2^{24}-1)\lambda<2^{127}$, so it is normal and finite in FP32 under IEEE~754~\cite{ieee2019754}. Every leaf product is representable, and induction over the kernel's additions, binary or fused, proves that every node is computed exactly. The prescribed conversion then gives $c_{32}$ or $c_{\mathrm{bf16}}$. \hfill$\square$

These are sufficient conditions, not necessary ones. Bounding only the final sum would be unsound, because cancellation can make it small after an earlier partial sum has already rounded. The kernel contract is a separate requirement from MX format compliance. It is stated for fused additions because tensor cores typically align several products and the accumulator to a common exponent and add them in one step inside an adder of limited width. Such an adder satisfies the contract only if it keeps at least 24 significant bits below the largest aligned exponent; otherwise an exactly representable sum can still lose low-order bits. We rely on this property and treat it as a per-kernel assumption to validate on each target platform.

\subsection{Sound Certificate Evaluation}
\label{sec:appendix-certificates}

To justify refinement, let $I_t$ be the $t$th 32-element block, and use the block statistics of Appendix~\ref{sec:impl-mechanism}. The block and tight bounds are
\begin{equation}
\begin{aligned}
B_{\mathrm{block}}&=\min\{\Lambda_1,\Lambda_2,\Lambda_3\},\qquad B_{\mathrm{tight}}=T,\\
\Lambda_1&=\textstyle\sum_t L_{a,t}M_{b,t},\qquad
\Lambda_2=\textstyle\sum_t M_{a,t}L_{b,t},\\
\Lambda_3&=\textstyle\sum_t\sqrt{Q_{a,t}Q_{b,t}}.
\end{aligned}
\label{eq:appendix-block-bound}
\end{equation}
Then $T=B_{\mathrm{tight}}\leq B_{\mathrm{block}}\leq B_{\mathrm{row}}$. To prove the first inequality, apply $\sum_{k\in I_t}|a_kb_k|\leq L_{a,t}M_{b,t}$, its symmetric counterpart, and Cauchy--Schwarz on each block, then sum. For the second, whole-row maxima dominate block maxima, and Cauchy--Schwarz across blocks gives $\sum_t\sqrt{Q_{a,t}Q_{b,t}}\leq\sqrt{Q_aQ_b}$. Each blockwise sum is bounded by its corresponding row expression; taking minima proves the result. Replacing the statistics of one operand by componentwise maxima over a row group gives an upper bound for every member. The exponent guards must hold for every member too.

\textbf{Rounding allowance.} Mathematical bounds become certificates only when their computed values remain upper bounds. Consider a dot product $S$ of nonnegative inputs, supplied exactly or enclosed upward, evaluated as $\widehat S$. Assume every rounding satisfies $\operatorname{fl}(q)\geq(1-\eta)q$ for its exact nonnegative input $q$, where $0\leq\eta<1$, with no underflow or overflow, as in the standard model of floating-point arithmetic~\cite{higham2002accuracy}. If each summand encounters at most $d$ roundings, including product formation and internal merges, expanding the reduction gives
\begin{equation}
\widehat S\geq(1-\eta)^dS,\qquad \overline S=\operatorname{RU}\bigl(\widehat S\,(1-\eta)^{-d}\bigr)\geq S.
\label{eq:appendix-inflation}
\end{equation}
Here $\operatorname{RU}$ denotes an upper enclosure of the entire expression. For normal results with $p$ significant bits, nearest rounding permits $\eta=2^{-p}$; truncation toward zero permits $\eta=2^{1-p}$.

Row statistics, square roots, and subsequent products must likewise be enclosed upward, using directed rounding or a justified error allowance, as in verified floating-point computation~\cite{muller2018handbook}. The row test can avoid the square root by testing an upper bound on $Q_aQ_b$ against $2^{48}$. Block tests apply Equation~\eqref{eq:appendix-inflation} to their nonnegative reductions. The minimum of independently sound upper bounds remains sound, although rounding allowances need not preserve the bounds' refinement order.

For the tight test, the exact physical sum is $\lambda T$, and every term $|x_ky_k|$ is nonnegative. This test needs one rounding allowance, a factor $1+2^{-7}$ for the BF16 conversion of the result. Suppose the kernel satisfies the contract of Proposition~1 and rounds inexact results monotonically, as round-to-nearest and truncation both do, and let $v$ be its computed result after BF16 conversion, if the kernel converts. If $T<2^{24}$, Proposition~1 applies to the sign-cleared operands under the guards of Equation~\eqref{eq:appendix-native-guards}, so the FP32 sum is $\lambda T$ exactly. Then $\lambda T\leq(1+2^{-7})v$, since BF16 keeps eight significant bits. If $T\geq2^{24}$, some partial sum reaches or exceeds $2^{24}\lambda$, which is a power of two and therefore representable in FP32 and in BF16; adding further nonnegative terms and rounding monotonically can never bring the computed value below it, so $v\geq2^{24}\lambda$. The test $(1+2^{-7})v<2^{24}\lambda$ therefore passes only if $T<2^{24}$, and when it passes, $T\leq\overline T=(1+2^{-7})v/\lambda<2^{24}$. The threshold $2^{24}\lambda$ is finite under the exponent guards and is formed by checked exponent arithmetic. Non-finite evaluations cannot certify an output.

\subsection{Exact Repair and Merging}
\label{sec:appendix-repair}

Repair must preserve information that native accumulation might lose, as in exact summation with a long accumulator~\cite{kulisch2013computer, collange2015numerical}. Partition a dot product into chunks with exact sums $s_r$, certified using their local units and guards. Express each chunk sum in the original row-product unit as $s_r=\lambda Z_r$, $Z_r\in\mathbb{Z}$. Nonzero chunks rebase by exact integer left shifts; zero chunks contribute zero. Sufficient capacity checks for arbitrary merge orders are
\begin{equation}
\sum_r|Z_r|<\begin{cases}2^{31}&\text{for signed int32},\\2^{53}&\text{for FP64}.\end{cases}
\label{eq:appendix-window}
\end{equation}
Every subset sum then fits exactly in the selected representation. With exact rebasing and checked scaling, induction over merges preserves $\lambda\sum_rZ_r=s$. Guarded FP64 accumulation of individual products follows the same argument with one product per chunk. A wider format alone is not a proof of exactness.

\textbf{Integer fallback.} Write each E2M1 element as $h_k/2$, where $h_k\in\mathbb{Z}$ and $|h_k|\leq12$. For a jointly nonzero block, let $d_t=\sum_{k\in I_t}h^x_kh^y_k$, and let $2^{\Delta_t}$ be the ratio of its block-product unit to the row-product unit $\lambda$. We have $|d_t|\leq4608$, and the E8M0 exponent range of the OCP MX specification~\cite{rouhani2023ocpmx} gives $0\leq\Delta_t\leq508$. Hence
\begin{equation}
\begin{aligned}
Z&=\sum_t d_t\,2^{\Delta_t},\qquad s=\lambda Z,\\
\sum_t|d_t|\,2^{\Delta_t}&\leq 4608\cdot\tfrac{K}{32}\cdot 2^{508}<2^{541}\quad\text{for }K/32<2^{20}.
\end{aligned}
\label{eq:appendix-integer-capacity}
\end{equation}
The sums range over jointly nonzero blocks. Under the checked condition $K/32<2^{20}$, 18 radix-$2^{32}$ limbs provide sufficient capacity, as in the superaccumulator of ExBLAS~\cite{iakymchuk2016exblas, collange2015numerical}. Splitting each shifted term into signed digits of magnitude below $2^{32}$ keeps every unpropagated limb below $2^{52}$ in magnitude; checked carry propagation fits int64. Finalization rounds $\lambda Z$ directly to FP32 using its retained bits, then to BF16 if requested. An intermediate conversion that discards low bits is not permitted. Invalid inputs, exceeded capacity, or unsupported output range require escalation or refusal.

\textbf{Proposition 2 (Partition invariance).} If disjoint sets $J_r$ cover a reduction of the same quantized operands, expressed in a common unit, exactly once, exact rebasing and merging of $Z_r=\sum_{k\in J_r}a_kb_k$ produce the same prescribed output for every partition and merge tree within checked capacity.

\emph{Proof.} Integer addition is associative and commutative, so every merge yields $\sum_rZ_r=\sum_ka_kb_k$. Exact rebasing preserves its value, and the same final conversion acts on the same $s$. \hfill$\square$

This is the reproducibility argument of order-free summation methods~\cite{collange2015numerical, ahrens2020algorithms}, restated for CertiSum's merges. Rounding local outputs before merging does not meet this premise. For example, FP32 rounding of $s_1=2^{24}+1$ and $s_2=-2^{24}$ before addition yields zero, whereas their exact sum is one. An ordinary FP32 collective is sufficient when a global certificate covers its intermediate sums.

\subsection{Attention and Training Invariance}
\label{sec:appendix-invariance}

To preserve the reduction terms across layouts, executions must retain the same logical inputs, parameters, optimizer state, masks, and random choices. They must preserve quantizer orientation, global 32-element blocks, backward $32\times32$ tiles, pruning decisions, global loss normalization, and rounding boundaries. Partition boundaries along every quantized dimension must fall on multiples of 32, so that no MX block straddles two partitions: the reduction dimension for tensor-parallel shards, the token dimension for micro-batches and data-parallel ranks, and the sequence dimension for context-parallel chunks. A block that straddled two sequentially computed micro-batches would need a scale that depends on data not yet available. Backward $32\times32$ tiles impose the same alignment on both of their dimensions. Numerical maps outside exact reductions, including exponentiation and quantization of attention weights, must agree across supported hardware. For attention this means that the exponentiation producing $w_{ij}$ returns identical quantized codes on every supported platform; Appendix~\ref{sec:appendix-exp-cross-platform} describes how MI350X and GB200 meet this requirement. FoldAttention~\cite{achanta2026foldattention} similarly fixes a softmax reference before the scan, which makes partial sums over separate key ranges additive.

The weight $w_{ij}$ is a function of the score $S_{ij}$, of the scale of its global key block, and of the per-row reference of Appendix~\ref{sec:impl-mechanism}, which is fixed by the global block positions of the first and the query-containing key blocks and does not depend on how keys are partitioned. The running maximum of blockwise softmax would not serve, since it depends on the visit order. Pruning is a function of the block scale and the same reference, so it is partition independent as well. Hence $w_{ij}$ and $\mathbf v_{ij}$ are fixed across layouts. The condition $\ell_i>0$ requires at least one unmasked, unpruned key per query row. Under a causal mask, the block that sets the reference contains an unmasked key, is never pruned, and has a nonzero largest quantized weight, so the condition holds; masks that can leave a row with no admissible key are outside the scope of this guarantee. For a row with $\ell_i>0$ and fixed $w_{ij}$ and $\mathbf v_{ij}$, a key partition $J_r$ contributes the exact pair
\begin{equation*}
(\mathbf n_{i,r},\ell_{i,r})=\Bigl(\sum_{j\in J_r}w_{ij}\mathbf v_{ij},\;\sum_{j\in J_r}w_{ij}\Bigr).
\end{equation*}
These dyadic sums admit exact integer representations. Proposition 2 applies componentwise, so merging the pairs recovers the same $\mathbf n_i$ and $\ell_i>0$. Finalization must round their exact quotient as in Equation~\eqref{eq:CertiSum-attention}. Let $m$ be the midpoint of adjacent FP32 candidates bracketing the exact quotient. The exact sign of $n_{i,d}-m\ell_i$ selects the side containing component $d$; ties use the even candidate, a standard technique for correctly rounded division~\cite{muller2018handbook}. Separately rounding either sum, or relying on an unchecked FP64 division, does not establish this contract.

\textbf{Corollary (Accepted training steps).} Under these conditions, suppose every partition-dependent reduction uses certified or repaired exact partial sums, exact merging, and the prescribed final conversion. This covers reductions over quantized GEMM and attention operands. A training step also contains reductions outside this class, such as the vocabulary-parallel log-sum-exp of cross-entropy, global gradient-norm clipping, and loss normalization. Each of them must either be merged exactly under Proposition~2 or be computed identically on every layout, for example from the same exactly merged inputs in a fixed order. Loss normalization in particular must be applied either as a power-of-two factor to the partial sums or once to the merged sum; applying a general factor to each partial sum before merging rounds before merging and violates the premise of Proposition~2. If other operators implement identical deterministic maps, every accepted step produces identical parameter and optimizer-state bits across layouts and supported hardware.

\emph{Proof.} Induct through the fixed logical computation graph. Equal predecessors produce identical quantized operands and local-map outputs. Proposition 2 gives equal reduction results; attention follows from the merged pair and common quotient rounding. Equal gradients, old parameters, and optimizer state give equal updates. Repeating this argument proves equality across accepted steps. Acceptance requires all reductions to be complete and all failures resolved across ranks; tentative values affected by a failed reduction must be recomputed before an update is committed. \hfill$\square$

\section{Experimental Details}
\label{sec:appendix-experimental}

This section collects setup details and caveats for the experiments in \S\ref{sec:evaluation}.

\subsection{Cross-Platform Invariance}
\label{sec:appendix-exp-cross-platform}

\textbf{TP4 configuration.} The TP4 configuration partitions computation while replicating parameters and optimizer state.

\textbf{Shared numerical definitions.} CertiSum and the vendor baseline use the same numerical definitions outside GEMMs and attention. RoPE and SiLU use shared lookup tables; loss and optimizer operations are explicitly rounded. Cross-entropy uses a shared correctly rounded exponential with a host fallback and host-evaluated logarithms. CertiSum attention fixes the MXFP4 rounding of $2^x$. GB200 uses threshold comparisons, while MI350X uses native \texttt{exp2} certified to produce the same quantized codes on reachable inputs.

\textbf{Layout invariance versus vendor agreement.} Each vendor's one-GPU and TP4 baseline losses match, so invariance across these layouts does not ensure cross-vendor agreement.

\subsection{Inference Invariance}
\label{sec:appendix-exp-inference}

\textbf{Decoding setup.} Each decoding step recomputes the full prefix, right-padded to 4,096 positions with padded keys and values zeroed.

\textbf{Repeatability versus invariance.} Repeatability alone does not ensure invariance. The twin's batch-dependent answer arises because its persistent GEMM changes summation order with the output row's position in the batch.

\subsection{Overhead}
\label{sec:appendix-exp-latency}

\textbf{Comparing implementations.} These are comparisons between implementations, since TP communication and operating clocks differ between CertiSum and its twin, and for the smaller CP cases the twin was not timed with graph replay. The latency gaps in \S\ref{sec:performance} include these implementation differences.

\textbf{Out-of-memory configurations.} Two further configurations, Qwen3-14B-Base on one GPU at $4\times4096$ and $1\times16384$, ran out of memory in every compared implementation.

\textbf{NVIDIA GB200.} Our GB200 experiments focus on platform invariance (\S\ref{sec:cross-platform}).

\section{General-Knowledge Questions}
\label{sec:appendix-questions}

The inference experiments in \S\ref{sec:inference} use the following 32 general-knowledge questions. They were generated with Claude Code and are reproduced verbatim in their original order.

\begin{generalquestions}
Why is the sky blue?
What causes the seasons on Earth?
How does a refrigerator keep food cold?
Why do we have leap years?
How does a vaccine work?
Why does ice float on water?
What is a prime number and why do prime numbers matter?
How does GPS determine a position?
Why do leaves change color in the autumn?
How do airplanes stay in the air?
What causes the tides in the ocean?
How does the human heart pump blood through the body?
Why does bread rise when it is baked?
How does a rainbow form?
What is photosynthesis and why is it important?
How do noise-cancelling headphones work?
Why do we dream?
How does a compass work?
What causes earthquakes?
How does the internet send data from one computer to another?
Why is the ocean salty?
How do bees make honey?
What is inflation in economics?
How does a microwave oven heat food?
Why do cats purr?
How does the immune system fight infections?
What causes thunder and lightning?
How do solar panels produce electricity?
Why do onions make people cry?
How does a credit score work?
What is the greenhouse effect?
How do volcanoes form?
\end{generalquestions}

\section{Training-Step Latency}
\label{sec:appendix-performance}

\textbf{AMD MI350X.} Table~\ref{tab:mi350x-training} lists the training-step latencies summarized in Figure~\ref{fig:performance} (\S\ref{sec:performance}). Each run performs 30 training steps; the first three are warm-up. We exclude a run if any timed step exceeds that step's median across runs of the same configuration by more than both 250\,ms and 25\%. Main shapes use 7--8 retained CertiSum runs, eight twin runs, and three runs per BF16 reference; smaller shapes use three runs per variant. Our implementations use pretrained weights and FineWeb-Edu; torchtitan uses random initialization, with FineWeb-Edu for Llama and a small C4 test corpus for Qwen. The static-route variant fixes routes during calibration and retains acceptance checks.

\begin{table*}[t]
\centering
\fontsize{9}{10}\selectfont
\setlength{\tabcolsep}{1.5pt}
\setlength{\parskip}{0pt}
\setlength{\aboverulesep}{1pt}
\setlength{\belowrulesep}{2pt}
\renewcommand{\arraystretch}{0.98}
\begin{tabular*}{\textwidth}{@{\extracolsep{\fill}}llrrrrr@{\hspace{7pt}}rrrrr@{\hspace{7pt}}rrrrr@{}}
\toprule
\multicolumn{2}{@{}l}{\textit{Main shapes}} & \multicolumn{5}{c}{Qwen3-8B-Base} & \multicolumn{5}{c}{Qwen3-14B-Base} & \multicolumn{5}{c}{Llama-3.1-8B} \\
\cmidrule(lr){3-7}\cmidrule(lr){8-12}\cmidrule(l){13-17}
Layout & $B\!\times\!S$ & $E$ (ms) & $E/T$ & $E/F$ & $D$ (ms) & $E/D$ & $E$ (ms) & $E/T$ & $E/F$ & $D$ (ms) & $E/D$ & $E$ (ms) & $E/T$ & $E/F$ & $D$ (ms) & $E/D$ \\
\midrule
1 GPU & $4\!\times\!4096$ & 823.1 & 1.57 & 0.88 & 1112.5 & 0.74 & \multicolumn{5}{c}{OOM} & 763.7 & 1.49 & 0.83 & 1079.0 & 0.71 \\
1 GPU & $1\!\times\!16384$ & 1042.7 & 1.46 & 0.88 & 1715.7 & 0.61 & \multicolumn{5}{c}{OOM} & 931.9 & 1.41 & 0.81 & 1587.4 & 0.59 \\
TP2 & $4\!\times\!4096$ & 816.4 & 1.20 & 0.90 & 976.9 & 0.84 & 1345.4 & 1.40 & 0.95 & 1512.5 & 0.89 & 749.2 & 1.23 & 0.90 & 895.9 & 0.84 \\
TP4 & $1\!\times\!16384$ & 562.3 & 1.12 & 0.88 & 825.1 & 0.68 & 992.3 & 1.38 & 1.00 & 1204.5 & 0.82 & 495.7 & 1.12 & 0.86 & 744.4 & 0.67 \\
CP2 & $1\!\times\!16384$ & 911.1 & 1.27 & 0.94 & 1137.5 & 0.80 & 1625.9 & 1.54 & 0.86 & 2058.1 & 0.79 & 823.9 & 1.24 & 0.90 & 1105.4 & 0.75 \\
CP4 & $1\!\times\!16384$ & 540.4 & 1.39 & 0.91 & 681.0 & 0.79 & 928.5 & 1.60 & 0.91 & 1123.6 & 0.83 & 475.6 & 1.33 & 0.83 & 648.2 & 0.73 \\
DP4 & $4\!\times\!4096$ & 1024.3 & 1.16 & 0.97 & 1298.3 & 0.79 & 1726.8 & 1.31 & 0.89 & 2261.7 & 0.76 & 928.8 & 1.12 & 0.89 & 1250.5 & 0.74 \\
\midrule
\multicolumn{17}{@{}l}{\textit{Smaller shapes}} \\
1 GPU & $1\!\times\!4096$ & 312.5 & 1.48 & 1.01 & 359.1 & 0.87 & 549.4 & 1.55 & 0.98 & 638.1 & 0.86 & 281.6 & 1.40 & 0.89 & 346.2 & 0.81 \\
TP2 & $1\!\times\!4096$ & 321.2\rlap{\textsuperscript{s}} & 1.43 & 1.13 & 310.8 & 1.03 & 597.9 & 1.77 & 1.35 & 467.3 & 1.28 & 301.7\rlap{\textsuperscript{s}} & 1.45 & 1.13 & 285.1 & 1.06 \\
TP4 & $1\!\times\!4096$ & 221.1\rlap{\textsuperscript{s}} & 1.28 & 1.05 & 214.7 & 1.03 & 412.3\rlap{\textsuperscript{s}} & 1.72 & 1.36 & 308.4 & 1.34 & 207.1\rlap{\textsuperscript{s}} & 1.38 & 1.12 & 190.5 & 1.09 \\
CP2 & $1\!\times\!8192$ & 516.5 & 1.20 & 0.80 & 690.4 & 0.75 & 961.8 & 1.52 & 0.74 & 1371.1 & 0.70 & 484.9 & 1.20 & 0.80 & 661.9 & 0.73 \\
CP4 & $1\!\times\!8192$ & 365.0\rlap{\textsuperscript{s}} & 1.15 & 0.84 & 451.3 & 0.81 & 589.8 & 1.44 & 0.82 & 753.0 & 0.78 & 376.1\rlap{\textsuperscript{s}} & 1.36 & 0.92 & 428.0 & 0.88 \\
DP4 & $1\!\times\!4096$ & 450.4 & 1.33 & 1.13 & 470.7 & 0.96 & 699.5 & 1.49 & 1.00 & 807.6 & 0.87 & 384.6 & 1.24 & 0.99 & 447.1 & 0.86 \\
\bottomrule
\end{tabular*}
\caption{\textbf{Training-step latency on AMD MI350X with FineWeb-Edu inputs.} Implementations $E$, $T$, $F$, and $D$ as in Table~\ref{tab:arms}. Ratios below one favor CertiSum. $B\!\times\!S$: sequences $\times$ tokens, global except DP4 (per GPU). \textsuperscript{s}Static-route $E$; unmarked $E$ and all $T$ are eager. OOM: all compared implementations ran out of memory before completing the first step.}
\label{tab:mi350x-training}
\end{table*}

\section{Routing and Repair}
\label{sec:routing}

To evaluate CertiSum's use of native and exact computation, we measure route allocation and recorded repairs. We instrument the four model/layout configurations in Table~\ref{tab:mi350x-routing} on FineWeb-Edu and The Stack v3 using AMD MI350X GPUs. Each configuration and dataset pair uses one 30-step counting run. We sum counters over all GPUs for the 27 steps after warm-up (indices 3--29).

Table~\ref{tab:mi350x-routing} separates work assigned to native GEMMs, direct exact kernels, and the recorded integer engine, as percentages of the step's nominal GEMM FLOPs. The forward-output columns cover transformer linear layers, excluding attention and the output head.\footnote{Nominal FLOPs count dense GEMMs as $2MNK$ and causal attention over valid pairs. Native candidates include outputs subsequently repaired; the repair fraction counts targets recorded by the instrumented repair paths. Recorded repair work sums $2K$ per counted repair dot product across forward and input-gradient paths.}

\begin{table}[!htb]
\centering
\fontsize{9}{10}\selectfont
\setlength{\tabcolsep}{1.5pt}
\setlength{\parskip}{0pt}
\setlength{\aboverulesep}{1pt}
\setlength{\belowrulesep}{2pt}
\renewcommand{\arraystretch}{1.02}
\begin{tabular*}{\columnwidth}{@{\extracolsep{\fill}}lrrrrrr@{}}
\toprule
& \multicolumn{3}{c}{Assigned work (\%)} & \multicolumn{2}{c}{Forward (\%)} & \multicolumn{1}{c}{Repair} \\
\cmidrule(lr){2-4}\cmidrule(lr){5-6}\cmidrule(l){7-7}
Dataset & Native & Exact & Int. & Native & Repair & Work (\%) \\
\midrule
\multicolumn{7}{@{}l}{\textit{Qwen3-8B-Base, 1 GPU, $4\!\times\!4096$}} \\
FineWeb-Edu & 36.92 & 63.08 & 0.0000 & 88.60 & 0.0229 & 0.0069 \\
The Stack v3 & 36.98 & 62.97 & 0.0451 & 88.89 & 0.0186 & 0.0072 \\
\addlinespace[3pt]
\multicolumn{7}{@{}l}{\textit{Qwen3-8B-Base, TP2, $4\!\times\!4096$}} \\
FineWeb-Edu & 37.19 & 62.81 & 0.0000 & 89.77 & 0.0558 & 0.0123 \\
The Stack v3 & 37.22 & 62.74 & 0.0376 & 89.91 & 0.0521 & 0.0111 \\
\addlinespace[3pt]
\multicolumn{7}{@{}l}{\textit{Qwen3-14B-Base, TP4, $4\!\times\!4096$}} \\
FineWeb-Edu & 54.23 & 45.76 & 0.0110 & 89.58 & 0.9710 & 0.2318 \\
The Stack v3 & 54.12 & 45.84 & 0.0349 & 89.15 & 0.8497 & 0.1945 \\
\addlinespace[3pt]
\multicolumn{7}{@{}l}{\textit{Llama-3.1-8B, CP4, $1\!\times\!16384$}} \\
FineWeb-Edu & 32.17 & 67.83 & 0.0005 & 98.81 & 0.1019 & 0.0189 \\
The Stack v3 & 32.16 & 67.84 & 0.0000 & 98.78 & 0.0829 & 0.0154 \\
\bottomrule
\end{tabular*}
\caption{\textbf{Routing and recorded repairs on AMD MI350X.} Shapes are global $B\!\times\!S$. Assigned work splits nominal step GEMM FLOPs among native, direct exact, and integer-engine (Int.) routes, the last including padding. Forward columns give native candidates as a share of transformer linear-layer outputs and recorded repairs as a share of those. Repair work is a share of nominal GEMM FLOPs. Values are rounded independently.}
\label{tab:mi350x-routing}
\end{table}

The native route handles 88.60--98.81\% of transformer linear-layer forward outputs. Instrumented paths record repairs for 0.0186--0.9710\% of native forward candidates.

Recorded repair work is 0.0069--0.2318\% of the step's nominal GEMM FLOPs. Qwen3-14B-Base TP4 has the largest recorded repair work, at 0.1945--0.2318\%; the two 8B models remain below 0.019\%. These counts quantify repair dot products, not all repair arithmetic or runtime overhead.

Exact MXFP4 kernels handle much of the full training step. Native GEMMs account for 32.16--54.23\% of assigned work, and direct exact kernels account for 45.76--67.84\%. Attention and nearly all weight-gradient work contribute to the exact-kernel share. Recorded integer-engine work is at most 0.0451\%, including padding.

We conclude that CertiSum uses the native route for most linear-layer forward outputs. Recorded repairs cover fewer than 1\% of native candidates. Direct exact kernels handle much of the full training step.

\end{document}